\documentclass[runningheads]{llncs}
\usepackage[T1]{fontenc}

\usepackage{graphicx}
\usepackage{url}
\usepackage{multirow}
\usepackage{float}

\begin{document}
\title{Machine Unlearning for the XGBoost Model with Network Intrusion Datasets}
\titlerunning{Machine Unlearning for XGBoost}  
%
%
\author{Diana Magalh{\~{a}}es\orcidID{0009-0007-1839-7164} \and
Eva Maia\orcidID{0000-0002-8075-531X} \and
Jo{\~{a}}o Vitorino\orcidID{0000-0002-4968-3653} \and
Isabel Pra{\c{c}}a\orcidID{0000-0002-2519-9859}}
\authorrunning{D. Magalh{\~{a}}es et al.}
%
\institute{GECAD, ISEP, Polytechnic of Porto, Rua Dr. António Bernardino de Almeida, 4249-015 Porto, Portugal\\
\email{\{dcpms,egm,jpmvo,icp\}@isep.ipp.pt}}
\maketitle              

\begin{abstract}
Machine Unlearning (MU) has emerged as an important technique for removing specific data points from trained models without requiring full retraining. However, most existing MU research focuses on deep learning and image data, leaving a gap in the domain of network intrusion detection, which relies heavily on tabular data. This work introduces XGBoost-Forget, an unlearning approach for the XGBoost model, to address this gap. The approach is evaluated on two tabular Network Intrusion (NI) datasets, IoT-23 and GeNIS, using multiple metrics to assess model performance, unlearning efficiency, and forgetting quality. The results show that XGBoost-Forget maintains predictive performance close to the original model while providing significantly faster unlearning, demonstrating its potential for MU in tabular NI settings.

\end{abstract}

\section{Introduction}
As organizations become increasingly interconnected, the number and complexity of malicious activities have increased, making cybersecurity a major concern for both industry and research. To defend against such risks, Network Intrusion Detection (NID) systems have been developed to monitor network traffic and identify patterns that could indicate malicious behavior \cite{b2}. Many of these systems rely on Artificial Intelligence (AI) and Machine Learning (ML) to detect both known and unknown threats, allowing them to respond more quickly and adaptively \cite{b2,b8}.

In this context, Network Intrusion (NI) datasets are essential for the development of NID systems. The quality of these datasets greatly influences how well ML models can distinguish between normal and malicious behavior \cite{b2}. Therefore, creating effective NID systems often relies on various benchmark datasets used for this purpose, including Internet of Things-23 (IoT-23) \cite{b3} and GECAD Network Intrusion Scenarios (GeNIS) \cite{b4}, which represent traffic from different types of network environments. However, NI datasets can contain data-quality issues that may require the removal of certain samples to maintain reliable model performance. For example, mislabeling \cite{b5}, which occurs when network traffic is assigned the wrong class, such as labeling benign traffic as malicious or vice versa. Since mislabeled data can negatively affect the model’s performance, it may need to be removed. While removing data from traditional databases may be simpler, removing it from ML models can be more complex. Even if the data is removed from the dataset, the trained model still reflects information learned from that sample, because the model has already been trained on it \cite{b38}.

Retraining ML models from scratch each time data must be deleted can be computationally expensive and inefficient. To address this challenge, the concept of Machine Unlearning (MU) has emerged. MU enables trained ML models to `forget' specific training data without requiring full retraining, effectively removing the impact of that information while preserving overall model performance \cite{b6}. Despite the growing interest in MU approaches, most existing research primarily focuses on their application to deep learning and image data \cite{b38,b39}. This indicates a significant gap in research regarding the application of unlearning approaches to traditional ML models and tabular data, a limitation that is especially relevant for NID, where network traffic is commonly represented in tabular datasets \cite{b8}.

Given the importance of tabular data and the strong performance of traditional models on this type of data, it is important to investigate how MU approaches can be adapted to both tabular datasets and traditional ML models. To address this gap, this work explores unlearning for the Extreme Gradient Boosting (XGBoost) model. The main contribution of this work is an unlearning approach for the XGBoost model, which, to the best of current knowledge, has not been previously explored. XGBoost was selected not only for its strong performance on tabular data but also because it has been shown to work effectively with NI datasets \cite{b9}.

The structure of this work is organized as follows. Section \ref{sec:related_work} reviews existing research on NID and MU, describing multiple unlearning approaches and discussing commonly used evaluation metrics. Section \ref{sec:methodology} details the design of the XGBoost-Forget approach, describing how it was inspired by the SISA framework to enable unlearning on the XGBoost model. Section \ref{sec:results} outlines the NID datasets used in the experiments and reports the results obtained using the selected metrics. Finally, Section \ref{sec:conclusions} summarizes the main findings of the work and reflects on their implications within the context of MU for tabular NID data.

\section{Related Work}
\label{sec:related_work}
Several studies applying ML to NID rely on traditional models, with tree-based methods appearing consistently across the literature \cite{b1}. For example, Nirupama and Niranjanamurthy developed a detection system using Decision Tree (DT) and Random Forest (RF), concluding that both models achieved high Accuracy (ACC) and a low false-alarm rate \cite{b24}. Rahman et al. compared the DT, logistic regression, RF, XGBoost, and support vector machine models, observing that tree-based methods provided high detection ACC \cite{b28}. Kiflay et al. introduced an ensemble approach combining RF, AdaBoost, Gradient Boosting, and XGBoost, which showed high ACC and reduced false alarms compared with individual classifiers \cite{b22}. Pandit et al. implemented ensemble combinations of DT, RF, Extra Trees, and XGBoost, reporting improved detection accuracy \cite{b25}. Araújo et al. used a stacking ensemble of XGBoost, LightGBM, and CatBoost, demonstrating that the integration of gradient-boosting models achieved high performance results \cite{b23}. Since tree-based models consistently demonstrate strong performance in NID, it is essential to explore how MU can be applied to these models.

\subsection{Machine Unlearning Approaches}
MU approaches are designed to enable a trained model to forget specific data samples or their influence without requiring complete retraining. Several methods have been proposed to achieve this objective, and existing research categorizes them into two main groups: Exact Unlearning and Approximate Unlearning.

Exact unlearning approaches completely remove a sample’s influence from a model \cite{b10}, ensuring full compliance with privacy regulations such as the GDPR’s `right to be forgotten'. Retraining the entire model from scratch after excluding the data to be forgotten is an example of exact unlearning. While this guarantees complete removal, it becomes impractical for large datasets due to the high computational cost of full retraining. To address this limitation, recent work explores exact unlearning approaches that avoid retraining the entire model while still eliminating all influence of the target data and preserving model integrity \cite{b11}.

One of the first exact unlearning approaches that helped avoid full retraining was SISA \cite{b12}. First, the training data is divided into multiple independent shards where each data point belongs to a single shard. Each shard is sliced to help reduce the retraining overhead. A model is trained in isolation for each shard, where the slices are presented incrementally, and the model’s parameter state is saved before each new slice is added, ensuring the data only influences each specific model. When an unlearning request arrives, only the model of the specific shard that contains the data to be unlearned needs to be retrained. Whereas SISA achieves unlearning by partitioning and isolating the data, other approaches modify the model itself to support efficient removal. DaRE \cite{b20} stores split statistics and instance information at nodes and leaves of the trees, allowing the model to use these cached statistics to identify which nodes are affected and retrain only the impacted subtrees, avoiding full retraining.

Approximate unlearning is another category of MU. The approaches under this category are designed to prioritize efficient unlearning, making them suitable for large-scale and complex models where exact unlearning may be computationally impractical \cite{b6}. Unlike exact unlearning, approximate unlearning focuses on speed rather than a complete removal guarantee. Its goal is to update the model efficiently while keeping its behavior close to that of a model trained without the deleted data, even if the removal is imperfect.

HedgeCut \cite{b18} introduces an unlearning approach for ensembles of randomized DT. It maintains robust and non-robust splits and computes alternative subtrees so that, when deletion requests occur, the model can simply adjust leaf statistics or switch subtrees without retraining. Lin et al. proposed an unlearning approach for Gradient Boosting Decision Trees (GBDTs) that modifies the training process to include random split selection and partitioning layers, reducing the number of subtrees affected by data removal. It also stores intermediate statistics to decide which parts of the model need to be updated, avoiding access to the original training data \cite{b19}. Similarly, DeltaBoost \cite{b21} redesigns the GBDT algorithm itself to make it more robust to deletions. It uses histogram-based splitting, bagging, and gradient quantization to minimize dependencies among trees, allowing efficient adjustment after data removal.

\subsection{Machine Unlearning Metrics}
As research on MU has progressed, multiple evaluation metrics have been introduced to assess the effectiveness of the proposed approaches. These metrics are typically grouped into three main categories: model utility, unlearning efficiency, and forgetting quality.

Model utility metrics measure how well the model preserves its predictive performance after unlearning, and they typically correspond to the standard metrics used in evaluating ML models. Accuracy (ACC), defined as the proportion of correctly classified predictions, is useful for determining whether the unlearning process has damaged the model’s predictive capability. Ideally, the unlearned model should retain ACC values close to those of the original model, even if a slight decrease happens \cite{b12,b18,b19,b20}. Recall (REC) is a metric used to evaluate the model’s ability to identify positive instances within the dataset accurately. Similar to ACC, the goal is for the unlearned model to maintain REC values close to those of the original model \cite{b32}. Precision (PRE) evaluates the proportion of correct positive predictions out of all positive predictions made by the model. Higher values indicate better performance in avoiding false positives. Similarly to the previous metrics, it is expected that the unlearned model achieves PRE values comparable to the ones obtained in the original model. The F1-Score (F1) is the harmonic mean between PRE and REC. It indicates the model’s ability to identify positive cases and minimize false ones. It is expected that the unlearned model achieves results comparable to those obtained by the original model \cite{b32}.

Unlearning efficiency metrics focus on how much time and computational effort it takes to remove the target data. Running Time (RT) evaluates the computational time required to run the unlearning process or retrain the model from scratch. It helps show whether an unlearning approach saves time compared to retraining. Ideally, the RT of the unlearning approach should be lower or equivalent to retraining, since one of the motivations for unlearning is avoiding the cost of retraining the whole model \cite{b18,b21}.

Forgetting quality metrics assess whether, and to what extent, the influence of the deleted data has been effectively removed from the model. Attack Success Rate (ASR) or Backdoor ACC is a metric used in experiments involving poisoning attacks \cite{b31}. A poisoning attack introduces a backdoor trigger into the training data, allowing the model to associate a specific hidden pattern with a target class. In the context of unlearning, ASR indicates whether the model continues to recognize the trigger introduced during training after the unlearning process has removed the poisoned samples. After unlearning, a lower ASR means that the unlearned model successfully forgets the poisoned samples present during training \cite{b36}. Kullback-Leibler Divergence (KLD) is a metric that measures the difference between two output distributions \cite{b33}. In the context of MU, it can be used to measure how closely the unlearned model’s output distribution aligns with the model retrained from scratch. Lower values indicate greater similarity between the unlearned and retrained models, reflecting more effective unlearning \cite{b36}. The Jensen–Shannon Divergence (JSD) is another metric that quantifies the difference between probability distributions and can be used to measure how closely the output distribution of the unlearned model aligns with that of a retrained model. However, unlike KLD, JSD is symmetric and bounded, which makes it easier to interpret. When a retrained model is not available, JSD can instead be computed between the original model and the unlearned model. In this case, a higher divergence is expected, since effective unlearning should make the unlearned model’s outputs differ from those of the original model. In the work by Perifanis et al., a retrained model was not available, so JSD was computed between the original model and the unlearned model. Although the analyzed study mentions using JSD, some of the reported results are listed as infinite, which should not happen with JSD since the metric is bounded. This suggests that the authors may not be using the true JSD metric \cite{b37}.

\section{Methodology}
\label{sec:methodology}
The objective of this work is to design and evaluate an unlearning approach for a tree-based model in the NID context. Since NID strongly relies on tabular data, models such as XGBoost have demonstrated superior performance, highlighting the need for an effective unlearning method tailored to XGBoost. For this purpose, the underlying principles of the SISA framework were adapted to the XGBoost model, leading to a SISA-inspired unlearning approach called XGBoost-Forget. In XGBoost-Forget, the training data is first divided into multiple shards, and each shard is used to train an independent XGBoost model. Within each shard, training is further split into slices, which are smaller portions of the shard's data. Across slices, the model is updated incrementally by adding new trees trained on additional data, and the model state after each slice is saved as a checkpoint. At inference time, predictions from all shard models are aggregated to produce the final output. When an unlearning request arrives, only the slices within the affected shard(s) that include the samples to be removed must be retrained. Slices that do not include the data to be removed can be reused from cache, while only the affected slices and those that follow are rebuilt using the remaining data. Finally, to demonstrate the efficiency and usability of the XGBoost-Forget approach in the NID context, two widely used NID datasets were selected, allowing it to be compared with the SISA framework under different traffic scenarios. Additionally, based on the related work, a set of evaluation metrics is used to compare the approaches in terms of performance and efficiency.

\section{Results}
\label{sec:results}
The datasets selected for this study were the IoT-23 and GeNIS datasets, both of which are publicly available and have been used for NID research. IoT-23 contains real IoT malware and benign traffic captures collected from 20 infected and 3 benign device scenarios. GeNIS includes a series of staged attack sequences along with multiple types of realistic benign traffic, generated during network simulations executed on the Airbus CyberRange platform. Table \ref{tab:datasets_distribution} summarizes each dataset's composition, including the total number of instances, how many were used for training and testing, the number of features, and the number of classes.

\begin{table}[h]
\vspace{-0.2cm}
\setlength{\tabcolsep}{3pt}
\caption{Datasets: Number of instances, features, and classes.}
\label{tab:datasets_distribution}
\centering
\begin{tabular}{|c|c|c|c|c|c|}
  \hline
  Dataset & Total Instances & Train Instances & Test Instances & Features & Classes \\
  \hline
  IoT-23 & 865,100 & 692,080 & 173,020 & 292 & 2 \\
  \hline
  GeNIS & 2,806,168 & 2,244,934 & 561,234 & 14 & 2 \\
  \hline
\end{tabular}
\end{table}

A series of experiments was conducted to demonstrate the capabilities of the XGBoost-Forget approach. To better assess its performance, the evaluation includes a comparison with the SISA framework implemented using a NN. For a fair baseline, the NN follows the architecture proposed by the SISA authors for their experiments on tabular data, which consists of two fully connected layers \cite{b12}. Experiments are conducted using configurations of 5 shards with 1 or 3 slices.

To accurately evaluate the efficiency of XGBoost-Forget, the instances to be unlearned were intentionally placed in a single shard. If these instances were spread across multiple shards, each affected shard would require retraining, which would not reflect the actual efficiency of the approach. This is relevant because the sharding strategy is most effective when only a small number of shards need to be retrained. Otherwise, the process starts to resemble full retraining. In total, 0.01\% of the training data was selected for unlearning in each dataset. For clarity throughout this section, a simplified naming scheme is adopted. XGB-T refers to XGBoost trained on the full dataset. XGB-R refers to XGBoost retrained without the forget set. XGB-FT denotes XGBoost-Forget after the initial training before unlearning, and XGB-FU represents XGBoost-Forget after the unlearning process. SISA corresponds to the SISA framework using a NN, after the unlearning process.

\subsection{IoT-23 Dataset}
Table \ref{tab:experiments_iot23_performance} summarizes the performance of all models with the IoT-23 dataset. It includes the baseline results for XGB-T and XGB-R, as well as the performance of XGB-FU and SISA when using 5 shards with either 1 or 3 slices. Examining the baselines, both XGB-T and XGB-R perform almost identically across all metrics, with only minor differences. This shows that removing the forget set did not significantly impact the model’s predictive performance. The RT of XGB-R serves as a reference point for evaluating the efficiency of XGB-FU.

When comparing XGB-FU to the baselines, it is noted that XGB-FU maintains a similar level of predictive performance, indicating that the unlearning process preserves the model’s accuracy. In terms of efficiency, XGB-FU is faster than full retraining across both slicing configurations. Relative to SISA, XGB-FU shows slightly higher predictive performance across all evaluation metrics, demonstrating the suitability of XGBoost for tabular datasets. In terms of RT, XGB-FU is also more efficient, completing the unlearning process faster than SISA under the evaluated configurations.

\begin{table}[h]
\setlength{\tabcolsep}{3pt}
\vspace{-0.2cm}
\caption{Performance for all models on the IoT-23 dataset.}
\label{tab:experiments_iot23_performance}
\centering
\begin{tabular}{|c|c|c|c|c|c|c|c|}
  \hline
  Shards & Slices & Model & ACC (\%) & PRE (\%) & REC (\%) & F1 (\%) & RT (s) \\
  \hline
  \multirow{2}{*}{-} & \multirow{2}{*}{-}
    & XGB-T & 98.172466 & 98.207847 & 98.172466 & 98.172466 & - \\ \cline{3-8}
    & & XGB-R & 98.175933 & 98.211741 & 98.175933 & 98.175621 & 5.2930 \\
  \hline
  \multirow{4}{*}{5} & \multirow{2}{*}{1}
    & XGB-FU & 98.251647 & 98.290905 & 98.251647 & 98.251318 & 2.1222 \\ \cline{3-8}
    & & SISA  & 92.264478 & 93.282266 & 92.264478 & 92.219410 & 18.5570 \\
  \cline{2-8}
    & \multirow{2}{*}{3}
    & XGB-FU & \textbf{98.286325} & \textbf{98.322732} & \textbf{98.286325} & \textbf{98.286028} & \textbf{1.3646} \\ \cline{3-8}
    & & SISA  & 92.262744 & 93.280962 & 92.262744 & 92.217645 & 8.4908 \\
  \hline
\end{tabular}
\end{table}

Table \ref{tab:experiments_iot23_asr} reports the ASR values for all evaluated configurations. It is important to mention that for this metric, the attack was implemented by selecting three features from the dataset and assigning each one a trigger value. This trigger was then inserted into the training instances meant to be unlearned, and the labels of those instances were set to class 0. For the test dataset, the same trigger was applied to all instances while keeping their original labels. 

\begin{table}[h!]
\setlength{\tabcolsep}{3pt}
\vspace{-0.2cm}
\caption{ASR results on the IoT-23 dataset.}
\label{tab:experiments_iot23_asr}
\centering
\begin{tabular}{|c|c|c|c|c|}
  \hline
  Scope & Shards & Slices & Model & ASR (\%) \\
  \hline

  \multirow{2}{*}{Retraining}
    & \multirow{2}{*}{-} & \multirow{2}{*}{-} & XGB-T & 99.9988 \\
  \cline{4-5}
    &                     &                     & XGB-R & 6.1218 \\
  \hline

  \multirow{4}{*}{All shards}
    & \multirow{4}{*}{5} & \multirow{2}{*}{1} & XGB-FT & 7.4477 \\
  \cline{4-5}
    &                     &                     & XGB-FU & \textbf{6.1692} \\
  \cline{3-5}
    &                     & \multirow{2}{*}{3} & XGB-FT & 8.1303 \\
  \cline{4-5}
    &                     &                     & XGB-FU & 7.7777 \\
  \hline

  \multirow{4}{*}{Infected shard only}
    & \multirow{4}{*}{5} & \multirow{2}{*}{1} & XGB-FT & 100.0000 \\
  \cline{4-5}
    &                     &                     & XGB-FU & \textbf{0.9779} \\
  \cline{3-5}
    &                     & \multirow{2}{*}{3} & XGB-FT & 99.9884 \\
  \cline{4-5}
    &                     &                     & XGB-FU & 10.4589 \\
  \hline

\end{tabular}
\end{table}

The table first shows results regarding the training and retraining baselines, which act as reference points for the rest of the ASR experiments using this dataset. Since retraining removes the poisoned samples entirely, it represents an ideal unlearning scenario. The table also includes the ASR results obtained when all shards are combined. As mentioned, only one of the five shards contains poisoned data, while the other four are clean. Because the final model prediction is produced by aggregating the outputs from all shards, the influence of the infected shard becomes diluted once all the data is combined. The dilution keeps the overall ASR low, even after unlearning. These observations suggest that the shard-based design may help mitigate the impact of backdoor triggers by limiting the influence of a single compromised shard on the final aggregated model. 

To accurately use ASR to measure forgetting quality, the table also reports the ASR values when only the infected shard is evaluated. All poisoned samples are inside a single shard, so that shard is where the backdoor behavior is stronger. Analyzing the results of the affected shard in isolation removes the dilution effect seen when all shards are combined, making the impact of unlearning easier to notice. In this scenario, ASR drops significantly after unlearning, with values closer to the ones obtained in the retraining baseline, indicating that forgetting is happening.

Table \ref{tab:experiments_iot23_jsd} shows that the observed JSD values are considerably lower than what would be expected for showing unlearning. Ideally, the divergence between XGB-T and XGB-FU should be closer to 1, indicating that unlearning significantly altered the model’s behavior, while the divergence between XGB-R and XGB-FU should be closer to 0. However, the values obtained are near zero in both cases. This suggests that JSD may not be well-suited for evaluating unlearning in this setting, possibly because the small removal ratio does not produce enough changes for the metric to capture.

\begin{table}[h]
\setlength{\tabcolsep}{3pt}
\vspace{-0.2cm}
\caption{Evaluation of JSD values for the IoT-23 dataset.}
\label{tab:experiments_iot23_jsd}
\centering
\begin{tabular}{|c|c|c|c|}
  \hline
  Slices & JSD(XGB-T, XGB-FU) & JSD(XGB-R, XGB-FU) \\
  \hline
  1 & 0.00022887 & 0.00043697 \\
  \hline
  3 &  0.00044204 &  0.00066481 \\
  \hline
\end{tabular}
\end{table}

\subsection{GeNIS Dataset}
Table \ref{tab:experiments_genis_performance} shows the performance of all evaluated models on the GeNIS dataset. For the baselines, XGB-T and XGB-R achieve nearly identical results across every metric. The RT of XGB-R, which is roughly four seconds, serves as the baseline for assessing the efficiency of XGB-FU.

When analyzing the unlearning results, XGB-FU maintains high predictive performance in both slicing configurations, matching the behavior seen in the baselines. This shows that the unlearning process does not negatively affect model performance when using a reasonable number of shards. In terms of efficiency, XGB-FU is faster than full retraining.

Compared to SISA, XGB-FU achieves slightly better predictive performance. However, XGB-FU completes the unlearning process faster than SISA, highlighting the efficiency of the XGB-FU approach. Overall, these results reinforce the suitability of XGBoost for tabular data and demonstrate that XGB-FU provides both accurate and efficient unlearning on the GeNIS dataset.

\begin{table}[h]
\setlength{\tabcolsep}{3pt}
\vspace{-0.2cm}
\caption{Performance for all models on the GeNIS dataset.}
\label{tab:experiments_genis_performance}
\centering
\begin{tabular}{|c|c|c|c|c|c|c|c|}
  \hline
  Shards & Slices & Model & ACC (\%) & PRE (\%) & REC (\%) & F1 (\%) & RT (s) \\
  \hline
  \multirow{2}{*}{-} & \multirow{2}{*}{-}
    & XGB-T & 99.978084 & 99.978057 & 99.978084 & 99.978031 & - \\ \cline{3-8}
    & & XGB-R & \textbf{99.978440} & \textbf{99.978414} & \textbf{99.978440} & \textbf{99.978389} & 3.9892 \\
  \hline

  \multirow{4}{*}{5} & \multirow{2}{*}{1}
    & XGB-FU & 99.977193 & 99.977170 & 99.977193 & 99.977129 & 0.7361 \\ \cline{3-8}
    & & SISA  & 99.207461 & 99.188691 & 99.207461 & 99.091624 & 34.5388 \\
  \cline{2-8}
    & \multirow{2}{*}{3}
    & XGB-FU & 99.978084 & 99.978065 & 99.978084 & 99.978023 & \textbf{0.5014} \\ \cline{3-8}
    & & SISA  & 99.206926 & 99.188132 & 99.206926 & 99.090902 & 17.5142 \\
  \hline
\end{tabular}
\end{table}

Table \ref{tab:experiments_genis_asr} presents the ASR values for the GeNIS dataset. As in the previous dataset, the retraining baseline shows a large ASR drop, confirming the expected behavior when the forget set is fully removed. This baseline acts as the reference point for evaluating the effectiveness of XGBoost-Forget on the GeNIS dataset.

The ASR values obtained when all shards are considered remain low both before and after unlearning. This matches what was observed in the IoT-23 dataset, since the final prediction aggregates the outputs of multiple shards, the influence of the single infected shard becomes diluted. These results reinforce the idea that the shard-based design limits the impact of backdoor triggers on the final aggregated model. When considering the infected shard results, the ASR drops to values close to zero in both slicing configurations. This decrease aligns with the retraining baseline, indicating that the unlearning procedure is successfully unlearning the infected samples.

\begin{table}[h]
\setlength{\tabcolsep}{3pt}
\vspace{-0.2cm}
\caption{ASR results on the GeNIS dataset.}
\label{tab:experiments_genis_asr}
\centering
\begin{tabular}{|c|c|c|c|c|}
  \hline
  Scope & Shards & Slices & Model & ASR (\%) \\
  \hline

  \multirow{2}{*}{Retraining}
    & \multirow{2}{*}{-} & \multirow{2}{*}{-} & XGB-T & 100.0000 \\
  \cline{4-5}
    &                     &                     & XGB-R & \textbf{0.0854} \\
  \hline

  \multirow{4}{*}{All shards}
    & \multirow{4}{*}{5} & \multirow{2}{*}{1} & XGB-FT & 1.5673 \\
  \cline{4-5}
    &                     &                     & XGB-FU & \textbf{1.5639} \\
  \cline{3-5}
    &                     & \multirow{2}{*}{3} & XGB-FT & 1.6177 \\
  \cline{4-5}
    &                     &                     & XGB-FU & 1.5797 \\
  \hline

  \multirow{4}{*}{Infected shard only}
    & \multirow{4}{*}{5} & \multirow{2}{*}{1} & XGB-FT & 100.0000 \\
  \cline{4-5}
    &                     &                     & XGB-FU & 1.5749 \\
  \cline{3-5}
    &                     & \multirow{2}{*}{3} & XGB-FT & 100.0000 \\
  \cline{4-5}
    &                     &                     & XGB-FU & 1.5840 \\
  \hline

\end{tabular}
\end{table}

Table \ref{tab:experiments_genis_jsd} presents the JSD values for the GeNIS dataset. Overall, the obtained values do not follow the expected pattern, aligning with the observations made for the IoT-23 dataset. These results suggest that the issue is likely not related to the data characteristics themselves but rather to the small removal ratio, which may not introduce enough changes for the JSD metric to reflect clear differences.

\vspace{-0.2cm}
\begin{table}[h]
\vspace{-0.2cm}
\setlength{\tabcolsep}{3pt}
\caption{Evaluation of JSD values for the GeNIS dataset.}
\label{tab:experiments_genis_jsd}
\centering
\begin{tabular}{|c|c|c|c|}
  \hline
  Slices & JSD(XGB-T, XGB-FU) & JSD(XGB-R, XGB-FU) \\
  \hline
  1 & 0.00000903 & 0.00000998 \\
  \hline
  3 & 0.00000696 & 0.00000795 \\
  \hline
\end{tabular}
\end{table}
\vspace{-0.2cm}

\subsection{Findings Discussion}
The experiments showed that, across all tested configurations, XGBoost-Forget maintains predictive performance very close to that of the original model. This indicates that the unlearning procedure, as evaluated, does not harm the model’s ability to make accurate predictions. In terms of efficiency, XGBoost-Forget consistently completes unlearning faster than full retraining. Overall, these results demonstrate that XGBoost-Forget can remove data while maintaining model quality and achieving good unlearning efficiency. Across both datasets, SISA reaches good performance, though XGBoost-Forget generally performs slightly better. The main difference is in the RT metric. XGBoost-Forget unlearns data faster than SISA in every configuration. This highlights the advantage of XGBoost-Forget for tabular data, offering high predictive ACC while providing more efficient unlearning.

Regarding forgetting-quality metrics, JSD did not behave as expected on either dataset. The consistently low values suggest that JSD may not be suitable in settings with small removal ratios, where the influence of the removed data is too small for the metric to capture meaningful changes in the distribution. Further investigation is needed to better understand the usefulness of JSD for evaluating unlearning in this case. ASR, on the other hand, proved to be a more informative metric for assessing forgetting quality. Across all configurations, ASR consistently dropped after unlearning and stayed close to the retraining baseline. The results obtained with this metric indicate that XGBoost-Forget is effectively removing the targeted data.

\section{Conclusions}
\label{sec:conclusions}
This work presented XGBoost-Forget, an unlearning approach that adapts the principles of the SISA framework to the XGBoost model. The goal was to explore MU in the context of traditional models and tabular NID data, an area that remains less explored compared to deep learning settings. The approach was evaluated using two NI datasets, IoT-23 and GeNIS, and compared with the original SISA framework.

The experimental results show that XGBoost-Forget maintains predictive performance close to the original model while providing significantly faster unlearning. These results suggest that the approach is well-suited for large and complex tabular datasets. The experiments also highlight the need to further explore unlearning evaluation metrics, as their behavior can vary depending on dataset characteristics, model structure, and the size of the removed data.

\section*{Acknowledgments}
   This research was funded by the Itea Project NADIR (\#22014) and Project NADIR (COMPETE2030-FEDER-01638900-19018) co-funded by Portugal 2030. Furthermore, this work also received funding from the project UID/00760/2025.

%
%

\bibliographystyle{splncs04}  
\bibliography{refs}

\end{document}